\documentclass[11pt,a4paper]{article}
\usepackage[hyperref]{acl} 
\usepackage{times}
\usepackage{latexsym}
\usepackage{url}
\usepackage{float}
\usepackage{natbib}

\usepackage{amsmath,amssymb,amsfonts}
\usepackage{algorithmic}

\usepackage{graphicx}
\usepackage{xcolor}

\usepackage[T1]{fontenc}
\usepackage[utf8]{inputenc}

\usepackage{hyperref}

\title{Comparative Study of Pre-Trained BERT and Large Language Models for Code-Mixed Named Entity Recognition}

\author{
    Mayur Shirke$^{1,3}$ \quad
    Amey Shembade$^{1,3}$ \quad
    Pavan Thorat$^{1,3}$ \quad
    Madhushri Wagh$^{1,3}$ \quad
    Raviraj Joshi$^{2,3}$ \\
    $^{1}$Dept. of Computer Engineering, Pune Institute of Computer Technology, Pune, India \\
    $^{2}$Indian Institute of Technology Madras, Chennai, India \\
    $^{3}$L3Cube Labs, Pune, Maharashtra, India \\
    \texttt{\small\{shirkemayur49, ameyshembade2511, thoratpavan64, madhushriwagh15\}@gmail.com} \\
    \texttt{\small ravirajoshi@gmail.com}
}

\date{}

\begin{document}
\maketitle

\begin{abstract}
Named Entity Recognition (NER) in code-mixed text, particularly Hindi-English (Hinglish), presents unique challenges due to informal structure, transliteration, and frequent language switching. This study conducts a comparative evaluation of code-mixed fine-tuned models and non-code-mixed multilingual models, along with zero-shot generative large language models (LLMs). Specifically, we evaluate HingBERT, HingMBERT, and HingRoBERTa (trained on code-mixed data), and BERT Base Cased, IndicBERT, RoBERTa and MuRIL (trained on non-code-mixed multilingual data). We also assess the performance of Google Gemini in a zero-shot setting using a modified version of the dataset with NER tags removed. All models are tested on a benchmark Hinglish NER dataset using Precision, Recall, and F1-score. Results show that code-mixed models, particularly HingRoBERTa and HingBERT-based fine-tuned models, outperform others — including closed-source LLMs like Google Gemini — due to domain-specific pretraining. Non-code-mixed models perform reasonably but show limited adaptability. Notably, Google Gemini exhibits competitive zero-shot performance, underlining the generalization strength of modern LLMs. This study provides key insights into the effectiveness of specialized versus generalized models for code-mixed NER tasks.
\end{abstract}

\section{Introduction}
In the field of Natural Language Processing (NLP), Named Entity Recognition (NER) plays a crucial role in extracting structured information from unstructured text. It involves identifying and classifying named entities such as persons, organizations, and locations within a sentence. NER has shown significant advancements in high-resource languages like English, largely due to the availability of large annotated corpora and the emergence of transformer-based models like BERT. However, these advancements do not directly translate to low-resource or linguistically diverse settings, such as code-mixed language data, where two or more languages are used interchangeably within the same sentence or conversation \cite{singh2018named,aguilar2018named}.

Code-mixing is a common linguistic phenomenon in multilingual societies such as India, where speakers often switch between languages like Hindi and English within the same utterance, resulting in informal and irregular sentence structures \cite{bali2014mixing}. Code-mixed data often includes transliteration, inconsistent grammar, and domain-specific vocabulary, making traditional NLP models less effective \cite{banerjee2017ner}. Moreover, standard benchmarks and pretrained models are not designed with code-mixed inputs in mind, limiting their ability to generalize and perform well in such scenarios \cite{khanuja2020gluecos}. This raises the need for models and methods that can understand and process code-mixed language effectively, especially for tasks like NER that rely heavily on syntactic and semantic cues.

To address these challenges, researchers have introduced transformer-based models that are fine-tuned specifically on code-mixed data \cite{nayak2022hingbert,chavan2023myboli}. Models such as HingBERT, HingMBERT, and HingRoBERTa have shown promise in handling Hindi-English code-mixed text due to their exposure to code-mixed corpora during pretraining or fine-tuning. These models adapt to the linguistic characteristics of mixed-language text and achieve improved performance over generic multilingual models in supervised learning settings. In particular, our results show that HingBERT-based fine-tuned models significantly outperform closed-source LLMs like Google Gemini for code-mixed NER, underscoring the importance of domain-specific pretraining. In contrast, widely used non-code-mixed models like BERT Base Cased, IndicBERT, RoBERTa, and MuRIL are trained on standard multilingual corpora and may lack exposure to the informal and domain-specific nature of code-mixed sentences.

In parallel, the emergence of large generative language models (LLMs) such as Google Gemini has opened up new avenues for zero-shot learning in NLP. These models, trained on vast and diverse datasets, have demonstrated impressive capabilities in understanding language across tasks and domains without task-specific fine-tuning \cite{team2023gemini}. While LLMs have primarily been evaluated on high-resource languages, their ability to generalize makes them a compelling candidate for tackling code-mixed NER in a zero-shot setting. However, the lack of annotated training data for such models demands careful dataset preprocessing—such as removing original entity labels—to create a realistic zero-shot evaluation setup.

This study aims to systematically compare the performance of three categories of models: (1) code-mixed fine-tuned models (HingBERT, HingMBERT, HingRoBERTa), (2) non-code-mixed multilingual models (BERT Base Cased, IndicBERT, RoBERTa and MuRIL), and (3) a zero-shot generative LLM (Google Gemini). By evaluating these models on a benchmark Hindi-English code-mixed NER dataset using metrics like Precision, Recall, and F1-score, we seek to understand the trade-offs between task-specific fine-tuning and generalization. Furthermore, we examine how generative models perform without access to annotated training data and explore their potential to replace or complement specialized models in resource-constrained settings.

The contributions and final observations of this work are as follows:
\begin{itemize}
    \item We present a comprehensive evaluation of code-mixed and non-code-mixed models for Hinglish NER, and analyze performance trends to highlight their strengths and limitations.  
    \item Code-mixed models such as \textbf{HingBERT} consistently outperform general-purpose models like standard BERT, demonstrating the importance of code-mixed pretraining.  
    \item While advanced LLMs (e.g., \textbf{Gemini}) show potential in zero-shot evaluation, they lag behind task-specific transformer models, indicating that domain adaptation remains crucial.  
\end{itemize}

Through this comparative study, we aim to guide future research in multilingual and cross-lingual NER and highlight promising directions for deploying LLMs in low-resource and linguistically diverse contexts.

\section{Related Work}
The task of Named Entity Recognition (NER) in code-mixed contexts, especially involving Hindi-English combinations, has emerged as a focal area in NLP due to the increasing prevalence of informal multilingual content across online platforms. The irregular grammar, spontaneous code-switching, and unconventional vocabulary present in such texts significantly hinder the performance of traditional NLP pipelines, which are typically designed for monolingual and well-structured language data.

Considerable efforts have been made in advancing NLP for Indian languages in code-mixed scenarios. For instance, the introduction of the MeCorpus—an extensive Marathi-English code-mixed corpus—has facilitated the development of tailored pre-trained models like MeBERT and MeRoBERTa. These models, along with annotated datasets for tasks like sentiment classification and hate speech detection, have established strong baselines that surpass generic multilingual models, thus laying a robust groundwork for Marathi-English code-mixed NLP research \cite{chavan2023myboli}.

Parallel efforts for Hindi-English include the development of HingCorpus, a large-scale dataset comprising over 52 million sentences, which was utilized to train models such as HingBERT and HingRoBERTa. These models have shown notable improvements across GLUECoS benchmark tasks, clearly demonstrating that leveraging real-world code-mixed data significantly enhances performance compared to models trained on synthetic or monolingual data \cite{nayak2022hingbert}.

To address the complexity of mixed-language texts, researchers have also proposed input augmentation techniques like word-level and sentence-level language tagging. Such augmentations, when applied without altering the base architecture, have consistently yielded better results in classification tasks related to sentiment, emotion, and hate speech \cite{takawane2023languageaug}.

Comparative analyses have reaffirmed the superiority of models trained on code-mixed data. For instance, HingBERT outperforms standard pre-trained models such as BERT, RoBERTa, and mBERT on various Hindi-English tasks, underscoring the necessity for domain-specific pretraining when dealing with multilingual and noisy text environments \cite{patil2023comparative}.

Earlier studies primarily explored the linguistic structure of Hindi-English code-mixed content. These investigations revealed that the breakdown of standard grammatical constructs in such texts poses major challenges for conventional NER methods \cite{bali2014mixing}. Follow-up work utilizing Conditional Random Fields (CRFs) for POS tagging emphasized the importance of preprocessing steps like token normalization and accurate language identification \cite{vyas2014pos}.

The CALCS shared task and related multilingual NER benchmarks have broadened research into other language pairs, such as Spanish-English and Arabic-English. In these efforts, systems utilizing multilingual embeddings showed clear advantages over rule-based approaches \cite{aguilar2018named}. Benchmarking platforms like GLUECoS have further enabled robust evaluation across code-switched language pairs, highlighting the utility of models like mBERT and XLM-R for capturing syntactic variability in mixed-language inputs \cite{khanuja2020gluecos}.

In terms of architecture, deep learning techniques like BiLSTM-CRF have gained traction for their effectiveness in capturing contextual and subword-level features, improving NER performance in noisy and informal language environments \cite{mayhew2017cheap}. With the rise of transformers, models trained or fine-tuned on mixed-language datasets have achieved significant gains, even in low-resource conditions \cite{winata2019learning}.

To broaden support for Indian language processing, several annotated code-mixed corpora have been released for other language pairs like Tamil-English and Malayalam-English \cite{chakravarthi2020corpus,chakravarthi2020sentiment}. These datasets promote the training and evaluation of NLP systems in broader multilingual contexts \cite{singh-etal-2018-named}.

The use of ensemble methods has also become increasingly prominent. Hybrid approaches that combine rule-based systems with statistical models such as CRFs have leveraged both syntactic and learned knowledge to improve accuracy \cite{jia2023review}. More advanced ensemble techniques incorporate multiple neural components—CNNs, BiLSTMs, transformers—to capture complementary linguistic signals \cite{liu2021hybrid}. Transformer ensemble systems have further used soft/hard voting and weighted averaging strategies across outputs from models like RoBERTa, mBERT, and XLM-R to boost robustness and reduce prediction variance \cite{ranasinghe2020multilingual}.

A notable example is the SemEval 2022 MultiCoNER shared task, where the CMNEROne team deployed a multilingual BERT-based architecture to effectively handle diverse code-mixed data \cite{dowlagar2022cmnerone}. Their success reinforces the growing relevance of multilingual pretraining for named entity recognition in complex linguistic environments.

Recent work has also explored new directions such as cross-script datasets and social media-specific entity taxonomies. One such effort introduced a Bengali-English dataset and tailored NER schema, producing promising results in named entity extraction for real-world applications like question answering \cite{banerjee2017ner}.

Beyond supervised methods, advances have also been made in handling noisy or incomplete annotations using distant supervision. For example, BOND-MoE applies a Mixture of Experts (MoE) strategy within an Expectation-Maximization framework to combine the outputs of multiple weak models, achieving strong performance even with noisy training data \cite{chen2024moe}.

Lastly, methods like Multilingual Meta-Embeddings (MME) have demonstrated how monolingual embeddings can be fused into unified multilingual representations without explicit language tags. This is achieved via self-attention mechanisms that learn to extract and integrate useful signals across languages. MME has shown strong generalization on code-switched NER tasks and highlights the potential of language-agnostic embedding fusion \cite{winata2019mme}.

\section{Methodology}
\subsection{Dataset Description}

The dataset employed in this study is a Hindi-English code-mixed corpus annotated for Named Entity Recognition (NER). It comprises 3,637 sentences, each associated with a unique sentence identifier, a list of tokenized words, and their corresponding NER tags in the BIO format. The tag set includes \texttt{B-Per} (Beginning of a Person entity), \texttt{I-Per} (Inside of a Person entity), \texttt{B-Org} (Beginning of an Organization entity), \texttt{I-Org} (Inside of an Organization entity), \texttt{B-Loc} (Beginning of a Location entity), \texttt{I-Loc} (Inside of a Location entity), and \texttt{Other} (non-entity or Other).

The overall distribution of entity tags is shown in Table~\ref{tab:label-distribution}. The dataset is inherently imbalanced, with a dominant number of \texttt{Other} tags, highlighting the sparsity of named entities in naturally occurring code-mixed text.

\begin{table}[h]
\centering
\caption{Label Distribution in the Dataset}
\label{tab:label-distribution}
\begin{tabular}{lr}
\hline
\textbf{NER Tag} & \textbf{Count} \\
\hline
B-Per & 2138 \\
I-Per & 554 \\
B-Org & 1432 \\
I-Org & 90 \\
B-Loc & 762 \\
I-Loc & 31 \\
Other     & 63497 \\
\hline
\end{tabular}
\end{table}

For model training and evaluation, we split the dataset into training and testing sets using an 80-20 ratio. This ensures that 80\% of the sentences are used for training while 20\% are reserved for evaluating model performance.

\subsection{Preprocessing}

To ensure consistency and effective model learning, several preprocessing steps are applied:

\begin{itemize}
    \item \textbf{Tokenization}: Each sentence is tokenized into subwords using the tokenizer associated with the respective pre-trained model. This step is crucial because modern transformer-based models operate on subword-level units.
    
    \item \textbf{Normalization}: Text is normalized to handle code-mixed artifacts such as spelling variations, inconsistent casing, and encoding differences. Unicode normalization and lowercasing are performed to reduce variance in textual forms.
    
    
    \item \textbf{Label Alignment}: Since tokenization may split words into multiple subwords, corresponding labels are aligned such that only the first subword token retains the original NER tag, while subsequent subwords are masked during loss computation to prevent label misalignment.
\end{itemize}

\subsection{Model Architectures}

To compare the performance of different modeling paradigms, we group our evaluated models into two categories:

\begin{itemize}
    \item \textbf{Non-Code-Mixed Models}: These include general-purpose multilingual or English-centric pre-trained transformers not specifically trained on code-mixed data. They serve as baselines for evaluating the effectiveness of code-mixed specific models.
    \begin{itemize}
        \item \textbf{BERT-base}: An English transformer model trained on BookCorpus and Wikipedia, serving as a foundational baseline.
        \item \textbf{RoBERTa-base}: A robustly optimized BERT variant trained on 160GB of English text using dynamic masking.
        \item \textbf{MuRIL}: A multilingual model designed for Indian languages, including Hindi-English, capturing diverse linguistic contexts.
        \item \textbf{IndicBERT}: A lightweight multilingual transformer supporting 12 Indian languages, optimized for efficiency in low-resource settings.
    \end{itemize}
    
    \item \textbf{Code-Mixed Specific Models}: These models are either fine-tuned or pre-trained on code-mixed datasets, making them more suitable for handling the linguistic irregularities of such text.
    \begin{itemize}
        \item \textbf{HingBERT}: A transformer model specifically fine-tuned on Hinglish data.
        \item \textbf{HingRoBERTa}: A RoBERTa variant adapted to Hindi-English code-mixed contexts.
        \item \textbf{Hing-mBERT}: A multilingual BERT model further trained on code-mixed data to enhance contextual understanding.
    \end{itemize}
\end{itemize}

\subsection{Fine-Tuning and Hyperparameter Optimization}

Each model is fine-tuned using the Hugging Face Transformers library, with hyperparameter search conducted via the Optuna framework. The search space includes learning rate, batch size, weight decay, Adam optimizer epsilon value, gradient accumulation steps, warmup steps, maximum gradient norm, and maximum input length. For each model, 40 optimization trials are performed, and the configuration yielding the highest validation F1-score is selected. Using these optimal hyperparameters, the models are retrained on the full training data, evaluated on the test set for comparative analysis, and saved in standard formats to ensure reproducibility and integration into downstream pipelines or ensemble systems.

\subsection{Evaluation Metrics}

We adopt the \texttt{seqeval} library for evaluation, which computes metrics at the entity level rather than the token level, providing a more accurate reflection of NER performance. The following metrics are reported:

\begin{itemize}
    \item \textbf{Precision}: The proportion of correctly predicted named entities to all predicted entities.
    \item \textbf{Recall}: The proportion of correctly predicted named entities to all actual entities in the ground truth.
    \item \textbf{F1-Score}: The harmonic mean of precision and recall, offering a balanced metric for performance.
    \item \textbf{Accuracy}: The overall correctness of predictions across tokens, excluding masked subwords.
\end{itemize}

These metrics are computed on the validation and test sets to facilitate both hyperparameter tuning and final model evaluation.

\subsection{Evaluation Using Large Language Model (LLM)}

To evaluate the performance of generative large language models in a zero-shot setting, we use the test and evaluation datasets by removing the original NER tags. These unlabeled datasets are then fed as input to the state-of-the-art LLM—Google's Gemini.

This model is prompted to perform NER by leveraging its internal knowledge and understanding of Hindi-English context. The predictions are post-processed and aligned with the original sentences, and compared to the ground truth labels using the same evaluation metrics.

This experiment provides insights into the feasibility of using LLMs for downstream tasks such as NER in code-mixed scenarios without additional fine-tuning.

\section{Experiments}

\subsection{Experimental Setup}

All experiments were conducted in the Kaggle Notebook environment, which provided two NVIDIA T4 GPUs (16 GB each), 32 GB RAM, and 4 CPUs. The software stack included Python 3.8, PyTorch 1.13, and the Hugging Face Transformers library (version 4.x). We used mixed-precision training (FP16) for efficiency, and computations were executed using CUDA for GPU acceleration.

\subsection{Implementation Details}

Each model was fine-tuned using the optimal hyperparameters identified via Optuna, as detailed in Section~3.4. A maximum sequence length of 128 tokens was used, and early stopping was applied based on validation F1-score with a patience of 5 epochs. 

Training was conducted using the AdamW optimizer with a linear learning rate scheduler and warm-up over the first 10\% of training steps. The batch size varied between 16 and 32 depending on the model size and available memory. To ensure reproducibility and robustness, each experiment was repeated with three different random seeds, and the average performance was reported.

\subsection{Evaluation Protocol and Discussion}

Each model was evaluated on the held-out 20\% test set using entity-level metrics from the \texttt{seqeval} library. We also analyzed model performance per entity type (PER, ORG, LOC) to understand differences in detection capabilities. All predictions were aligned to the original token boundaries, masking subword extensions appropriately. 

For large language models (LLMs), a zero-shot evaluation was performed. Gemini Pro was prompted with sentences (without labels) and asked to identify named entities in BIO format. Its predictions were post-processed and compared against ground truth. 

The results of these experiments are presented in Section~\ref{sec:results}, comparing model performance, highlighting the benefits of code-mixed specific pretraining, and examining the feasibility of zero-shot NER using large language models in resource-constrained environments.

\section{Results}

The evaluation results for all models are summarized in Table~\ref{table:eval_results} (validation set) and Table~\ref{table:test_results} (test set). These tables report Accuracy, Precision, Recall, and F1-scores across three groups of models: general-purpose BERT variants, code-mixed BERT models, and large language models (LLMs).  

\begin{table}[H]
\centering
\resizebox{\linewidth}{!}{%
\begin{tabular}{|l|c|c|c|c|}
\hline
\textbf{Model} & \textbf{Accuracy} & \textbf{Precision} & \textbf{Recall} & \textbf{F1-Score} \\
\hline
\multicolumn{5}{|c|}{\textbf{BERT Models}} \\
\hline
BERT Base Cased & 96.88 & 69.31 & 72.85 & 71.04 \\
IndicBERT       & 97.11 & \textbf{75.81} & 75.63 & 75.72 \\
RoBERTa         & 97.12 & 72.60 & 75.63 & 74.09 \\
MuRIL           & 97.22 & 72.35 & 74.70 & 73.51 \\
\hline
\multicolumn{5}{|c|}{\textbf{Code-Mixed BERT Models}} \\
\hline
HingBERT        & 97.33 & 72.89 & 80.51 & 76.51 \\
HingRoBERTa     & 97.36 & 74.50 & 77.95 & 76.19 \\
HingMBERT       & \textbf{97.43} & 73.27 & \textbf{81.43} & \textbf{77.14} \\
\hline
\multicolumn{5}{|c|}{\textbf{LLMs}} \\
\hline
Google Gemini   & 95.19 & 71.02 & 59.13 & 64.11 \\
\hline
\end{tabular}%
}
\caption{Evaluation Metrics on Validation Set (After Hyperparameter Tuning)}
\label{table:eval_results}
\end{table}

\begin{table}[H]
\centering
\resizebox{\linewidth}{!}{%
\begin{tabular}{|l|c|c|c|c|}
\hline
\textbf{Model} & \textbf{Accuracy} & \textbf{Precision} & \textbf{Recall} & \textbf{F1-Score} \\
\hline
\multicolumn{5}{|c|}{\textbf{BERT Models}} \\
\hline
BERT Base Cased & 96.58 & 70.77 & 73.53 & 72.12 \\
IndicBERT       & 97.07 & 78.66 & 74.40 & 76.47 \\
RoBERTa         & 96.98 & 75.91 & 76.57 & 76.24 \\
MuRIL           & 97.05 & 76.92 & 78.09 & 77.50 \\
\hline
\multicolumn{5}{|c|}{\textbf{Code-Mixed BERT Models}} \\
\hline
HingBERT        & \textbf{97.38} & \textbf{78.81} & 80.69 & 79.74 \\
HingRoBERTa     & 97.21 & 78.47 & 78.30 & 78.39 \\
HingMBERT       & 97.15 & 75.80 & \textbf{82.21} & \textbf{78.87} \\
\hline
\multicolumn{5}{|c|}{\textbf{LLMs}} \\
\hline
Google Gemini   & 95.52 & 67.16 & 58.04 & 62.24 \\
\hline
\end{tabular}%
}
\caption{Evaluation Metrics on Test Set (After Hyperparameter Tuning)}
\label{table:test_results}
\end{table}

\label{sec:results}
The experimental results highlight the comparative effectiveness of different categories of pre-trained models on the Hindi-English code-mixed Named Entity Recognition (NER) task. We evaluate the models on both validation and test sets, along with a separate zero-shot setup using Google Gemini.

The code-mixed fine-tuned models—HingBERT, HingMBERT, and HingRoBERTa—consistently outperform the other models across both validation and test datasets. Among them, HingMBERT achieves the highest F1-score, indicating strong generalization and robustness. These results confirm the advantages of task-specific pretraining on real code-mixed data, as these models are better equipped to handle informal sentence structures, transliterations, and frequent code-switching.

In contrast, the non-code-mixed multilingual models such as mBERT, IndicBERT, and MuRIL perform reasonably well, especially on metrics like accuracy and recall. However, their lower precision and F1-scores suggest that they struggle to identify named entities in inconsistent or informal language patterns. This emphasizes the limitation of models trained solely on formal multilingual corpora when applied to noisy code-mixed data.

The zero-shot evaluation of Google Gemini demonstrates its capacity to generalize without any task-specific fine-tuning. Despite not being trained explicitly for NER or on code-mixed inputs, Gemini achieves a relatively high overall accuracy. However, its low macro-averaged precision and F1-score reveal that the model favors frequent classes and struggles with fine-grained entity classification in the absence of supervision. This indicates that while generative large language models hold promise, they still face challenges in class-balanced performance on structured prediction tasks like NER.

Overall, these findings illustrate a clear performance gap between code-mixed and non-code-mixed models, validating the importance of domain-specific pretraining. They also highlight the emerging potential of large generative models in zero-shot learning, especially for low-resource language settings, though further improvements in structured prediction are required.

\section{Conclusion}
This paper presents a comparative study of various pre-trained models for Named Entity Recognition in Hindi-English code-mixed text. We evaluated code-mixed fine-tuned models, non-code-mixed multilingual models, and a zero-shot generative large language model (Google Gemini) on a standard benchmark dataset. The study highlights the importance of domain-specific training for handling code-mixed inputs and explores the potential of generative models for zero-shot NER. Our results show that generic large language models are still far behind fine-tuned models like HingBERT in handling code-mixed NER tasks effectively.

Our work establishes a foundation for further research in multilingual and code-mixed NER, particularly in low-resource settings. Future directions include exploring ensemble methods, extending evaluations to other language pairs, and improving generalization through cross-lingual pretraining and augmentation techniques.

This study also emphasizes the need for more robust benchmarks and evaluation protocols tailored specifically for code-mixed language scenarios. As code-switching becomes increasingly common in real-world communication, addressing its challenges remains critical for building inclusive and effective NLP systems.

\section*{Acknowledgments}
This work was done under the L3Cube Labs, Pune mentorship program. We want to thank our mentors at L3Cube for their continuous support and encouragement. 

\section*{Limitations}
Our study is limited by the relatively small dataset size, reliance on standard evaluation metrics, and lack of domain-specific or instruction-tuned model exploration. Moreover, fine-tuning large transformer models requires significant computational resources, which may hinder wider applicability.

\bibliography{main.bib}

\end{document}